\documentclass[final]{cvpr}

\usepackage{times}
\usepackage{epsfig}
\usepackage{graphicx}
\usepackage{amsmath}
\usepackage{amssymb}

\usepackage{multirow}
\usepackage{bm}
\usepackage{comment}

\usepackage{subfigure}
\usepackage{algorithm}
\usepackage{algorithmic}

\usepackage{pifont}

\newcommand{\cmark}{\ding{51}}%
\newcommand{\xmark}{\ding{55}}%

\usepackage[pagebackref=true,breaklinks=true,colorlinks,bookmarks=false]{hyperref}

\def\cvprPaperID{1043} 
\def\confYear{CVPR 2021}

\begin{document}

	\renewcommand{\algorithmicrequire}{\textbf{Input:}} 
	\renewcommand{\algorithmicensure}{\textbf{Output:}} 

	\newcommand{\fracpartial}[2]{\frac{\partial #1}{\partial  #2}}
	\newcommand{\norm}[1]{\left\lVert#1\right\rVert}
	\newcommand{\innerproduct}[2]{\left\langle#1, #2\right\rangle}
	\newcommand{\fan}[1]{\Vert #1 \Vert}
	\newcommand{\qileft}{[\kern-0.15em[}
	\newcommand{\qiLeft}{\left[\kern-0.4em\left[}
	\newcommand{\qiright}{]\kern-0.15em]}
	\newcommand{\qiRight}{\right]\kern-0.4em\right]}
	\newcommand{\sign}{{\mbox{sign}}}
	\newcommand{\diag}{{\mbox{diag}}}
	\newcommand{\armin}{{\mbox{argmin}}}
	\newcommand{\rank}{{\mbox{rank}}}
	\newcommand{\lbar}{\left\|}
	\newcommand{\rbar}{\right\|}

	\renewcommand{\a}{{\bm{a}}}
	\newcommand{\e}{{\bm{e}}}
	\newcommand{\f}{{\mathbf{f}}}
	\newcommand{\g}{{\bm{g}}}
	\newcommand{\m}{{\bm{m}}}
	\newcommand{\p}{{\bm{p}}}
	\newcommand{\s}{{\bm{s}}}
	\newcommand{\w}{{\bm{w}}}
	\newcommand{\x}{{\bm{x}}}
	\newcommand{\y}{{\bm{y}}}
	\newcommand{\z}{{\bm{z}}}
	\newcommand{\balpha}{{\bm{\alpha}}}
	\newcommand{\bmu}{{\bm{\mu}}}
	\newcommand{\bsigma}{{\bm{\sigma}}}
	\newcommand{\blambda}{{\bm{\lambda}}}
	\newcommand{\bgamma}{{\bm{\gamma}}}
	\newcommand{\btheta}{{\bm{\theta}}}
	\newcommand{\bbeta}{{\bm{\beta}}}
	
	\newcommand{\bpi}{{\bm{\pi}}}
	
	\newcommand{\ba}{{\bm{A}}}
	\newcommand{\bb}{{\bm{B}}}
	\newcommand{\bc}{{\bm{C}}}
	\newcommand{\bd}{{\bm{D}}}
	\newcommand{\be}{{\bm{E}}}
	\newcommand{\bg}{{\bm{G}}}
	\newcommand{\bi}{{\bm{I}}}
	\newcommand{\bj}{{\bm{J}}}
	\newcommand{\bl}{{\bm{L}}}
	\newcommand{\bp}{{\bm{P}}}
	\newcommand{\bq}{{\bm{Q}}}
	\newcommand{\bs}{{\bm{S}}}
	\newcommand{\bu}{{\bm{U}}}
	\newcommand{\bv}{{\bm{V}}}
	\newcommand{\bw}{{\bm{W}}}
	\newcommand{\bx}{{\bm{X}}}
	\newcommand{\by}{{\bm{Y}}}
	\newcommand{\bz}{{\bm{Z}}}
	\newcommand{\bTheta}{{\bm{\Theta}}}
	\newcommand{\bSigma}{{\bm{\Sigma}}}
	
	\newcommand{\A}{{\mathcal{A}}}
	\newcommand{\B}{\mathcal{B}}
	\newcommand{\C}{\mathcal{C}}
	\newcommand{\D}{\mathcal{D}}
	\newcommand{\E}{\mathcal{E}}
	\newcommand{\F}{\mathcal{F}}
	\newcommand{\G}{\mathcal{G}}

	\newcommand{\I}{\mathcal{I}}
	\newcommand{\K}{\mathcal{K}}
	\renewcommand{\L}{\mathcal{L}}
	\newcommand{\N}{\mathbb{N}}
	\newcommand{\W}{\mathcal{W}}
	\newcommand{\X}{\mathcal{X}}
	\newcommand{\Y}{\mathcal{Y}}
	\newcommand{\Q}{\mathcal{Q}}
	\newcommand{\R}{\mathbb{R}}
	
	\newcommand{\T}{\mathcal{T}}

	\newcommand{\1}{\mathbbm{1}}
	
	\newcommand\independent{\protect\mathpalette{\protect\independenT}{\perp}}
	
	\newcommand{\indep}{\rotatebox[origin=c]{90}{$\models$}}
	
	\newcommand{\aka}{\textit{aka}}

	\title{Manifold Regularized Dynamic Network Pruning}
	
	\author{Yehui Tang$^{1,2}$, Yunhe Wang$^{2}$\thanks{Corresponding author.}, Yixing Xu$^{2}$, Yiping Deng$^{3}$, Chao Xu$^{1}$,  Dacheng Tao$^{4}$, Chang Xu$^{4}$ \\
		\normalsize$^1$ Key Lab of Machine Perception (MOE), Dept. of Machine Intelligence, Peking University.
		\\
		\normalsize$^2$ Noah's Ark Lab, Huawei Technologies. \normalsize$^3$ Central Software Institution, Huawei Technologies.\\
		 \normalsize$^4$ School of Computer Science, Faculty of Engineering, University of Sydney.\\
		\small\texttt{yhtang@pku.edu.cn;~yunhe.wang@huawei.com;~c.xu@sydney.edu.au}
	}

	\maketitle
 	

\begin{abstract}
	
Neural network pruning is an essential approach for reducing the computational complexity of deep models so that they can be well deployed on resource-limited devices. Compared with conventional methods, the recently developed dynamic pruning methods determine redundant filters variant to each input instance which achieves higher acceleration. Most of the existing methods discover effective  sub-networks for each instance independently and do not utilize  the relationship between different inputs. To maximally excavate redundancy in the given network architecture, this paper proposes a new paradigm that dynamically removes redundant filters by embedding the manifold information of all instances into the space of pruned networks (dubbed as ManiDP). We first investigate the recognition complexity and feature similarity between images in the training set. Then, the manifold relationship between instances and the pruned sub-networks will be aligned in the training procedure. The effectiveness of the proposed method is verified on several benchmarks, which shows  better performance in terms of both accuracy and computational cost compared to the state-of-the-art methods. For example, our method can reduce 55.3\% FLOPs of ResNet-34 with only 0.57\% top-1 accuracy degradation on ImageNet.

\end{abstract}

\section{Introduction}

Deep convolutional neural networks (CNNs) have achieved  state-of-the-art performance on a large variety of computer vision tasks, \eg, image classification~\cite{he2016deep,tang2020semi,krizhevsky2017imagenet,you2020greedynas}, objection detection~\cite{girshick2015fast,redmon2016you,guo2020hit}, and video analysis~\cite{feichtenhofer2017spatiotemporal,guo2019beyond,jiang2018deepvs}. Besides the model performance, recent researches pay more attention on the model efficiency, especially the computational complexity~\cite{yang2020cars,su2021locally,singh2020leveraging}. Since there are considerable real-world applications required to be deployed on resource constrained hardwares, \eg, mobile phones and wearable devices, techniques that effectively reduce the cost of modern deep networks are required~\cite{lin2019towards,han2020ghostnet,you2020shiftaddnet,you2017learning}.

To this end, a number of model compression algorithms have been developed without affecting network performance. For instance, quantization~\cite{wu2016quantized,han2020training,rastegari2016xnor,yang2020searching} uses less bits to represent network weights and knowledge distillation~\cite{hinton2015distilling,xu2019positive,chen2020learning,park2019relational,xu2020kernel} is to train a compact network based on the knowledge of a teacher network. Low-rank approximation~\cite{lu2017fully,lebedev2014speeding,lin2018holistic} tries to decompose the original filters to smaller ones while pruning method directly discards the redundant neurons to get a sparser network. Among them, channel  pruning (or filter pruning)~\cite{he2017channel,tang2020scop, luo2018thinet,liu2017learning} is regarded as a kind of structured pruning method, which directly discards redundant filters to obtain a compact network with lower computational cost. Since the pruned network can be well employed on mainstream hardwares to obtain considerable speed-up, channel pruning is widely used in industrial products.

The conventional channel pruning methods obtain a static network applied to all input samples, which do not  excavate redundancy maximally, as  the diverse demands for network parameters and capacity from different instances are neglected. 
In fact,  the importance of filters is  highly input-dependent. A few methods proposed recently  prune channels according to individual instances dynamically and achieve better performance. For example, Gao~\etal~\cite{gao2018dynamic} introduce small auxiliary modules to predict the saliencies of channels with given input data, and prune unimportant filters at run-time. Instance-wise sparsity is adopted in~\cite{ijcai2019-416} to induce different sub-networks for different samples. However, the existing methods  prune channels for individual instances independently, which neglects the relationship  between different instances. A sparsity constraint with same intensity  is usually used for different input instances, regardless of the diversity of instance complexity. Besides, the similarity between instances is also valuable information deserving to explore.

\begin{figure*}
	\centering
	\centering
	\includegraphics[width=1.8\columnwidth]{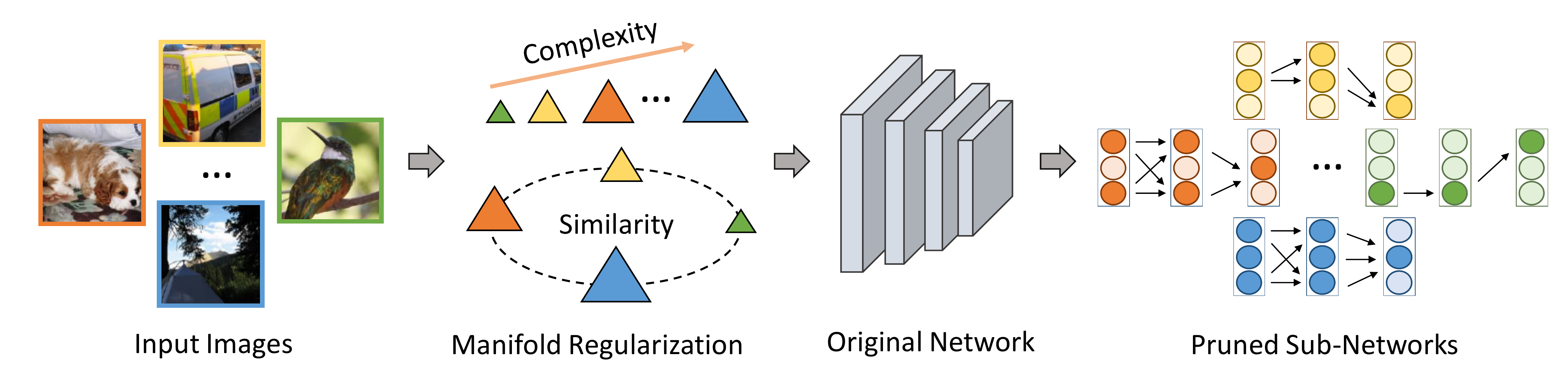} %
	\caption{Diagram of the proposed manifold regularized dynamic pruning method (ManiDP). We first investigate the complexity and similarity of images in the training dataset to excavate the manifold information. Then, the network is pruned dynamically by exploiting the manifold regularization.}
	\label{fig-pipeline} 
	\vspace{-5mm}
\end{figure*}

In this paper, we explore a new paradigm for dynamic pruning to maximally excavate network redundancy corresponding to arbitrary instance. The manifold information of all samples in the given dataset is exploited in the training process and  corresponding sub-networks are derived to preserve the relationship between different instances~(Figure~\ref{fig-pipeline}).  Specifically, we first propose to identify the complexity of each instances in the training set and adaptively adjust the penalty weight on channel sparsity. Then, we further preserve the similarity between samples in the pruned results, \ie, the sub-network for each input sample. In practice, the features with abundant semantic information obtained by the network are used for calculating the similarity. By exploiting the proposed approach, we can allocate the overall resources more reasonably, and then obtain pruned networks with higher performance and lower costs. Experiments are throughly conducted on a series of benchmarks for demonstrating the effectiveness of the new method. Compared with the state-of-the-art pruning algorithms, we obtain higher performance in terms of both network accuracy and speed-up ratios.

The rest of this paper is organized as follows: Section~\ref{sec-rel} briefly reviews the existing channel pruning methods and Section~\ref{sec-pre} introduces the formulations. We discuss the proposed method in Section~\ref{sec-approach} and conduct extensive experiments  in Section~\ref{sec-exp}. Finally, Section~\ref{sec-con} summarizes the conclusions.

\section{Related Work} 
\label{sec-rel}

Channel Pruning is a kind of coarse-grain structural pruning method that discards the whole redundant filters to obtain a compact network, which can achieve practical acceleration without specific hardware~\cite{wen2016learning,tang2020reborn,lin2018accelerating,liu2017learning,chen2020frequency}. It contains the conventional static pruning methods and recent dynamic algorithms, and we briefly review them  as follows.

\textbf{Static Pruning.} A compact network shared by different instances is desired in static pruning. Wen~\etal~\cite{wen2016learning} impose structural sparsity on the weights of convolutional filters to discover and prune redundant channels. Liu~\etal~\cite{liu2017learning} associates a scaling factor to each channel and the sparsity regularization is imposed on the factors. Recently, more methods are proposed  which achieve state-of-the-art performance on several benchmarks. For example, Molchanov~\etal~\cite{molchanov2019importance} uses Taylor expansion to estimate the contribution of a filter to the final output and discard filters with small scores, while Liebenwein~\etal~\cite{Liebenwein2020Provable} construct an importance distribution that reflects the filter importance.  To reduce the disturbance of  irrelevant factors, Tang~\etal~\cite{tang2020scop} set up a scientific control during pruning filters, which can discover compact networks with high performance. These methods prune same filters for different input instances and obtain a  'static' network with limited representation capability, whose performance degrades obviously when  a large pruning rate is required.

\textbf{Dynamic Pruning.} 
Beyond the static pruning methods, an alternative way is to determine the importance of filters according to input data, and skip unnecessary calculation in the test phase~\cite{rao2018runtime}. Dong~\etal~\cite{dong2017more} use low-cost collaborative layers to  induce sparsity on the original convolutional kernels at the running time.
Hua~\etal~\cite{hua2019channel} generate decision maps by partial input channels to identify the unimportant  regions in feature maps. However, the skipped ineffective regions in \cite{hua2019channel} are irregular and practical acceleration depends on special hardware such as FPGAs and ASICs.
Gao~\etal~\cite{gao2018dynamic} introduces squeeze-excitation modules to predict the saliency of channels and skip those with less contribution to the classification results.   Complementary to them, this paper focuses on  effectively training the dynamic network to allocate a proper sub-network for each instance, which is vital to achieve a satisfactory trade-off between accuracy and computational cost.

\section{Preliminaries} 
\label{sec-pre}
 In this section, we introduce the formulations of channel pruning for deep neural networks and the dynamic pruning problem.

Denote 	the dataset with $N$ samples as $\X=\{\x_i\}_{i=1}^N$, and $\Y=\{\y_i\}_{i=1}^N$ are the corresponding labels.
For a  CNN model with $L$ layers,  $W^l \in \R^{c^{l}\times c^{l-1} \times k^l \times k^l }$ denotes  weight parameters of the convolution filters in the $l$-th layer. $F^l(\x_i)\in \R^{c^l \times w^l \times h^l}$ is the output feature map of the $l$-th layer with $c^l$ channels, which can be calculated with convolution filters $W^l$ and the features $F^{l-1}(\x_i)$ in the previous layer, \ie, $F^{l}(\x_i)=ReLU(F^{l-1}(\x_i) \ast W^l)$, where $\ast$  denotes the convolutional operation and $ReLU(t)=\max(t,0)$ is the activation function.

Channel pruning  discovers and eliminates redundant channels in a given neural network to reduce the overall computational complexity while retaining a comparable performance~\cite{wen2016learning,liu2017learning,molchanov2019importance,Liebenwein2020Provable}. Basically, the conventional channel pruning can be formulated as
\begin{equation}
	\label{eq-cp}
	\min_{\W} \sum_{i=1}^N \L_{ce}(\x_i,\W)+ \lambda \cdot \sum_{l=1}^L \|W^l\|_{2,1}
\end{equation}
where $\W$ denotes all the weight parameters of the network, and  $\L_{ce}(\x_i,\W)$ is the task-dependent loss function (\eg, the cross-entropy loss for classification task).
$\|\cdot \|_{2,1}$ is the $\ell_{21}$-norm that induces channel-wise sparsity on convolution filters, \ie, $\|W^l \|_{2,1}=\sum_{j=1}^{c^l}\|vector(W^l_{j,:,:,:})\|_2$, where $vector(\cdot)$ straightens the tensor $W^l_{j,:,:,:}$ to vector form and  $\|\cdot\|_2$ is the $\ell_2$-norm\footnote{Note that the methods using scaling factors can also be regarded as this form as the factors can be absorbed to convolution kernels~\cite{liu2017learning}}. The trade-off coefficient $\lambda$ balances the two losses, and a larger $\lambda$ induces sparser convolution kernel $W^l$ and then obtain  a more compact network. 

Dynamic pruning is developed for further excavating the correction between input  instances and pruned channels~\cite{gao2018dynamic,rao2018runtime,hua2019channel}. Wherein, the importance of output channels depends on the inputs, and different sub-networks will be generated for each individual instance for flexibly discovering the network redundancy. To this end,
a control module $\G^l$ is introduced to process the input feature $\F^{l-1}(\x_i)$  and predict the channel saliency $\bpi^l(\x_i,\W)=\G^l(F^{l-1}(\x_i)) \in \R^{c^l}$ in the $l$-th layer for input $\x_i$.\footnote{In the following, we denote channel saliency $\bpi^l(\x_i,\W)$ as $\bpi^l(\x_i)$ for brevity.} In practice, a smaller element in $\bpi^l(\x_i)$ implies that the corresponding channel is less important.  The controller  $\G$ is usually implemented by utilizing a squeeze-excitation module  as suggested in~\cite{gao2018dynamic}. Then, redundant channels are determined through a gate operation, \ie, $\hat \bpi^l(\x_i)=\I(\bpi^l(\x_i),\xi^l)$, 
 where the element $\I(\bpi^l(\x_i),\xi^l)[j]$ is set to 0 when $\bpi^l(\x_i)[j]$ is less than the threshold $\xi^l$ and keeps unchanged otherwise.
By exploiting the mask $\hat \bpi^{l-1}(\x_i)$ in the previous layer, the feature $\F^{l}(\x_i)$ can be efficiently calculated as:
\begin{equation}
	\label{eq-dy1} 
	F^{l}(\x_i)= ReLU \left( \left (F^{l-1}(\x_i) \odot \hat \bpi^{l-1}(\x_i) \right) \ast W^l\right), 
\end{equation}
where $\odot$ denotes that each channel of feature $F^{l-1}(\x_i)$ is multiplied by the corresponding element in mask $\hat \bpi^{l-1}(\x_i)$. Since $\hat \bpi^{l-1}(\x_i)$ is usually very sparse and the calculation of redundant channels are skipped, the computational complexity of Eq.~(\ref{eq-dy1}) will be significantly lower than that of the original convolution layer. 

To retain the desirable performance, the dynamic network is also trained with both the cross-entropy loss  $\L_{ce}(\x_i,\W)$ and the sparsity regularization, \ie,
\begin{equation}
	\label{eq-dyp}
	\min_{\W} \sum_{i=1}^N \L_{ce}(\x_i,\W)+ \lambda \cdot \sum_{l=1}^L \|\bpi^{l}(\x_i)\|_1,
\end{equation} 
where $\|\bpi^{l}(\x_i)\|_1$ is the $\ell_1$-norm penalty on the channel saliency $\bpi^{l}(\x_i)$.  Obviously, a larger coefficient $\lambda$ also produces sparer saliency $\bpi^{l}(\x_i)$ and thus yields more compact networks with lower computational cost. Compared to the static pruning method~(Eq.~(\ref{eq-cp})) that uses a same compact network to handle all the input data, the dynamic one~(Eq.~(\ref{eq-dyp})) aims to prune  channels  for different instances accordingly.

\section{Manifold Regularized Dynamic Pruning}
\label{sec-approach}

The main purpose of dynamic pruning is to fully excavate the network redundancy for each instance. However, the manifold information, \ie, the relationship between samples in the entire dataset has rarely been studied. The manifold hypothesis states that the high-dimensional data can be embedded into low-dimensional manifold, and samples locate closely in the manifold space own analogous properties. The mapping function from input samples to their corresponding sub-networks should be smooth over the manifold space, and then the relationship between samples need to be preserved in sub-networks. This manifold information can effectively regularize the solution space of instance-network pairs to help allocate a proper sub-network for each instance. In the following, we explore the manifold information from two complementary perspectives, \ie, complexity and similarity.

\subsection{Instance Complexity}
\label{sec-adp}

For a given task, the difficulty of accurately predicting example labels   can be  various, which thus implies the necessity of investigating models with different capacities for different inputs. 
Intuitively, a more complex sample with vague semantic information~(\eg, images with insignificant objects, mussy background, \etc) may need a more complex network with a strong representation ability to extract the effective information, while  a  much simpler network lower computational cost could be enough to make the correct prediction for a simpler instance. Actually, this intuition reflects the relationship between instances on a 1-dimensional  complexity space, where the different instances are sorted according to their difficulties for the given task. 
To exploit this property, we firstly measure the complexity of instances and sub-networks, respectively, and then develop an adaptive objective function to align the complexity relationship between instances and that between sub-networks.

Considering that input instances are expected to have correct predictions made by the networks, the task-specific loss~(cross-entropy loss) $\L_{ce}(\x_i,\W)$ of the networks is adopted to measure the complexity of current input $\x_i$.  A larger cross-entropy loss implies that the current instance has not been fitted well, which is more complex and needs a  network with stronger representation capability for extracting the information. For a sub-network, the sparsity of channel saliency $\bpi^l(\x_i)$ determines the number of effective filters in it, and a sparser $\bpi^l(\x_i)$ induce a more compact network  with lower complexity. Hence, we use the sparsity of channel saliencies as the measurement of network complexity.

\begin{figure}[t]
	\centering
	\centering
	\includegraphics[width=0.8\columnwidth]{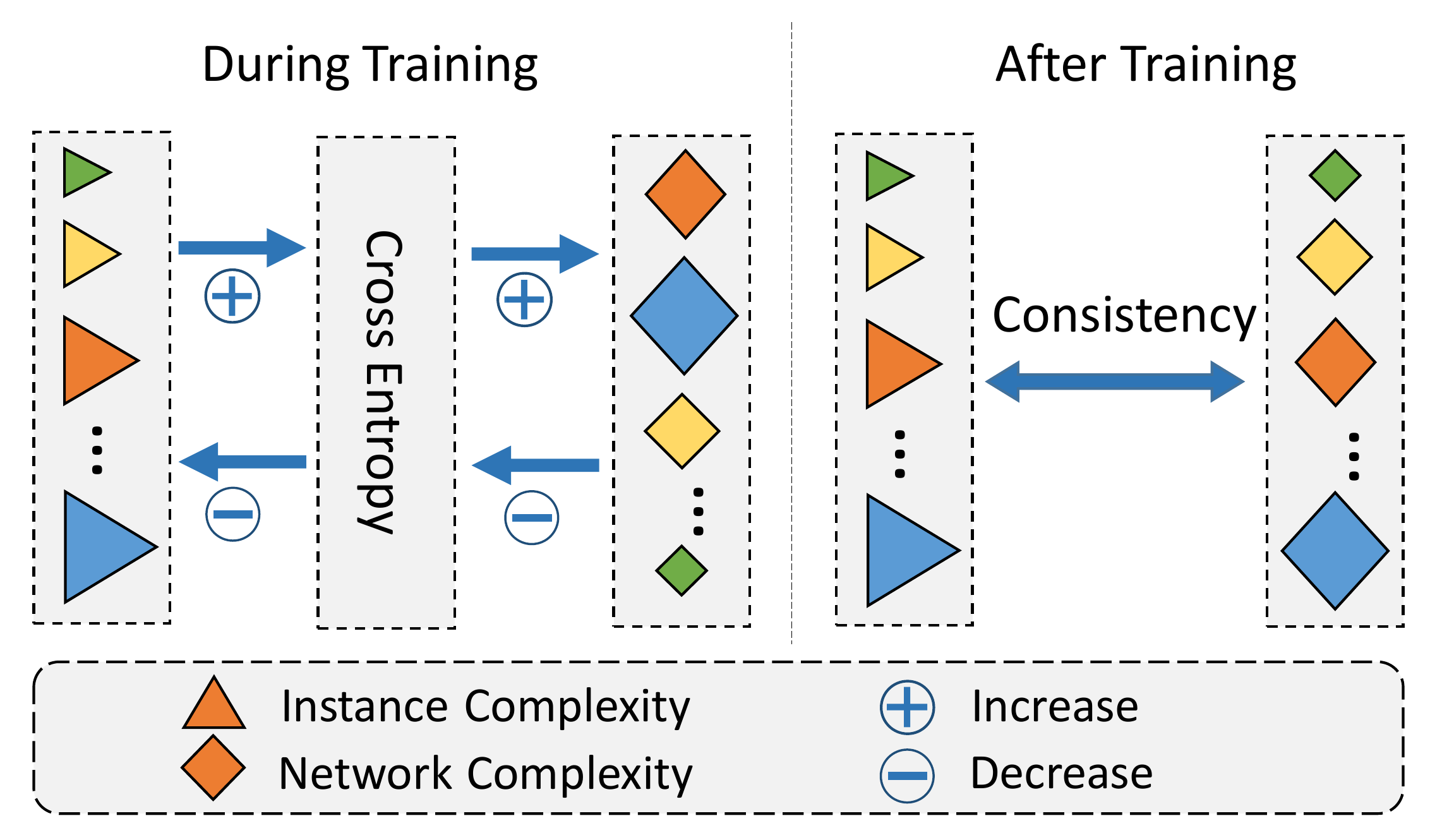} %
	\caption{The aligning process of instance complexity and network complexity. During training, the complexity of instances and networks will be automatically adjusted, and achieve consistency after training.}
	\label{fig-comp} 
	\vspace{-5mm}
\end{figure}

Recall that in Eq.~(\ref{eq-dyp}), the weight coefficient $\lambda$ is vital to determine the strength of sparsity penalty on channel saliencies. However, a same weight coefficient is assigned to all different instances without discrimination. In fact, a simple instance  whose cross-entropy loss can be minimized easily may desire a compact sub-network  for  computational efficiency. On the other hand, those examples that have not been well fitted by the current network yet would need more network capacity to pursue the prediction accuracy, instead of pushing the sparsity further. Thus, the weight of sparsity  penalty that controls the network complexity should increase when the cross-entropy loss  decreases and vice versa. In an extreme case, no sparsity constraint should be given to the corresponding sub-networks for those under-fitted examples. Specifically, a set of binary learnable variables $\bbeta =\{\beta_i\}_{i=1}^N\in \{0,1\}^N$ are used to indicate whether the sub-network for input instance $\x_i$ should be thinned out. Thus the optimization objective can be formulated as:
\begin{equation}
	\vspace{-2mm}
	\small
	\label{eq-lasso2} 
	\begin{aligned} 
		 \max_{\bbeta} \min_{\W} & \sum_{i=1}^N \L_{ce}(\x_i,\W) \\
		&+ \lambda' \cdot  \beta_i 
		\frac{C-\L_{ce}(\x_i,\W)}{C} \sum_{l=1}^L \|\bpi^{l}(\x_i)\|_1,
	\end{aligned}
\vspace{-0.5mm}
\end{equation} 
where $\lambda'$ is   a hyper-parameter shared by all instances to balance the classification accuracy and network sparsity. $\|\cdot\|_1$ is the $\ell_1$-norm that induces sparsity on channel saliencies and $\W$ is the set of network parameters. $C$ is a pre-defined  threshold for instance complexity and the samples with cross-entropy losses larger than $C$ are considered as over complex. Eq.~(\ref{eq-lasso2}) is optimized in a  min-max paradigm, \ie, the minimization is applied on parameter $\W$ to train the network, while the maximization on variables $\bbeta$ indicates whether the corresponding sub-networks should be made sparse. In practice,  $\bbeta$ has a closed-form solution. Note that $\|\bpi^{l}(\x_i)\|_1\ge0$  always holds, and  the optimal solution of $\bbeta$ only depends on the relative magnitude of cross-entropy loss $\L_{ce}(\x_i,\W)$ and the complexity threshold $C$, \ie,
\begin{equation}
	\vspace{-1.5mm}
	\label{eq-beta}
	\beta_i= \left\{
	\begin{array}{rcl} 
		1, & & \L_{ce}(\x_i,\W) \le C,\\
		0, & & \L_{ce}(\x_i,\W) > C.
	\end{array}\right.
	\vspace{-1mm}
\end{equation}
Eq~(\ref{eq-beta}) indicates that no sparsity is imposed on the corresponding sub-network ($\beta_i=0$) if the cross-entropy loss $\L_{ce}(\x_i,\W)$ of instance $\x_i$ exceeds $C$. In practice, the network is trained in mini-batch, and we empirically use the average cross-entropy loss over the whole dataset in the previous epoch as the threshold $C$. For brevity, the coefficient for the sparsity loss is denoted as $\lambda(\x_i)$, \ie,
\begin{equation}
	\lambda(\x_i)=\lambda' \cdot \beta_i\frac{C-\L_{ce}(\x_i,\W)}{C}.
\end{equation}
The value of $\lambda(\x_i)$ is always in range [0, $\lambda'$] and it has a larger value for simpler instances.
Then, the max-min optimization problem~(Eq.~(\ref{eq-lasso2})) can be simplified as:
\begin{equation}
	\label{eq-lasso3}
	\min_{\W} \sum_{i=1}^N \L_{ce}(\x_i,\W) 
	+ \lambda(\x_i)\cdot \sum_{l=1}^L \|\bpi^{l}(\x_i)\|_1.
\end{equation}
In the training process mentioned above, a negative feedback mechanism~\cite{acuto2008tailoring, belloc2008deadenylation} naturally exists to dynamically control the instance complexity and network complexity. As shown in Figure~\ref{fig-comp},
when an instance $\x_i$ is sent to the network and produces a large cross-entropy loss $\L_{ce}(\x_i,\W)$, it is considered as a complex instance and the penalty weight $\lambda(\x_i)$ is reduced to induce a complex sub-network. On account of the powerful representation capability of the complex network, the cross-entropy loss of the same instance $x_i$ can be easily minimized, and reduce the relative complexity of the instance. If the dynamic network takes a simple instance as input, the dynamic process is just the opposite. This negative feedback mechanism stabilizes the training process, and finally sub-networks with appropriate model capability for making correct prediction are allocated to the input instances.

\subsection{Instance Similarity}
\label{sec-cts}

Besides mapping  instances to the complexity space, the similarity between samples is also an effective clue to  customize the networks for different instances.
Inspired by the manifold regularization~\cite{wang2004adaptive,zhu2018image}, we expect that the  instance similarity  can be well preserved by their corresponding sub-networks, \ie, if two instances are similar, the allocated sub-networks for them tend to own similar property as well.  

The intermediate features $F^l(\x_i)$ produced by a deep neural network can  be treated as an effective representation of input sample $\x_i$. 
Compared to the original data, ground-truth information is embedded to the intermediate features during training, and hence the intermediate features are more suitable to measure the similarity between different samples. 
The  sub-network  for instance $\x_i$ can be described by the channel saliencies, which determine the architecture of the network. 
Note that both the channel saliencies $\bpi^l(\x_i)$ and intermediate features $F^l(\x_i)$ are corresponding to each layer, and thus we can calculate the similarity matrix of channel saliencies $T^l\in\R^{N\times N}$ and similarity matrix of features $R^l\in\R^{N\times N}$ layer-by-layer, where the element $T^l[i,j]$~($R^l[i,j]$) in the matrix reflect the similarity between saliencies (features) derived from different sample $\x_i$ and $\x_j$.  
Suppose that the classical cosine similarity~\cite{nguyen2010cosine} is adopted, $T^l$ is calculated as:
\begin{equation}
	T^l[i,j]=	 \frac{ \bpi^l(\x_i) \cdot \bpi^l(\x_j)}{\|\bpi^l(\x_i)\|_2\cdot \|\bpi^l(\x_j)\|_2},
\end{equation} 
where $\|\cdot\|_2$ denotes  $\ell_2$-norm. 
For intermediate  feature $F^l(\x_i) \in \R^{c^l \times w^l \times h^l}$ in the $l$-the layer, it is first flattened to a vector using the average pooling operation $p(\cdot)$,  and then the similarity matrix is:
\begin{equation}
	R^l[i,j]=	 \frac{ p(F^l(\x_i)) \cdot p(F^l(\x_j))}{\|p(F^l(\x_i))\|_2\cdot \|p(F^l(\x_j))\|_2}. 
\end{equation}
Given the similarity matrices $T^l$ and $R^l$ mentioned above, a loss function $\L_{sim}$ is developed to impose consistency constraint on them, \ie, 
\begin{equation}
	\label{eq-cons}
	\L_{sim}(\X,\W)=\sum_{l=1}^{L}dis(T^l,R^l),
\end{equation}
where $\X=\{\x_i\}_{i=1}^N$ and $\W$ denote the input data and network parameters, respectively, and  $dis(\cdot,\cdot)$  measures the difference between the two similarity matrices. Here we adopt a simple way that compares the corresponding elements of the two matrices, \ie, $dis(T^l,R^l)=\|T^l-R^l\|_F$, where $\|\cdot\|_F$ denotes Frobenius norm. In the practical implementation of network training, the similarity matrices are calculated over input data in each mini-batch for efficiency. 

Combining  the adaptive sparsity loss~(Eq.~(\ref{eq-lasso3})), the final objective function for training the dynamic network is:
\begin{equation} 
	\label{eq-final}
	\begin{aligned}
	\min_{\W} \sum_{i=1}^{N} & \L_{ce}( \x_i,\W) + \lambda(\x_i)\cdot\sum_{l=1}^L \|\bpi^{l}(\x_i)\|_1 \\
	 &+ \gamma \cdot \L_{sim}(\X,\W),
	 \end{aligned}
\end{equation}
where $\gamma$ is a weight coefficient for the consistency loss $\L_{sim}(\X,\W)$. In Eq.~(\ref{eq-final}), the manifold information is simultaneously excavated from two  complementary perspectives, \ie, complexity  and similarity. The former imposes the consistency between instances complexity and sub-networks complexity, while the latter induces instances with similar features to select similar sub-networks. Though different perspectives are emphasized, both loss functions describe intrinsic relationships between instances and networks, and can be simultaneously optimized to get an optimal solution.

Based on the channel saliencies of each layer, the dynamic pruning is applied to the given network for different input data separately. Given $N$ different input examples $\{\x_i\}_{i=1}^N$, the average channel saliencies   over different instance are first calculated as
$\bar \bpi^l=\{\bar \bpi^l[1], \bar \bpi^l[2], \cdots, \bar \bpi^l[c^l]\}$, where $c^l$ is the number of channels in the $l$-th layer. Then the elements in $\bar \bpi^l$ is sorted so that $\bar \bpi^l[1] \le \bar \bpi^l[2] \le \cdots \le \bar \bpi^l[c^l]$ and the threshold is set as $\xi^l=\bpi^l[\lceil\eta c^l\rceil]$, where $\eta$ is the pre-defined  pruning rate and $\lceil\cdot\rceil$ denotes round-off. At inference,  only channels with saliencies larger than the threshold $\xi^l$ need to be calculated and the redundant features are skipped, which reduces the computation and memory cost. Based on the threshold $\xi^l$ derived from the average saliencies $\bar \bpi^l$,  the actual pruning rate are different for each instances, since the channel saliency $\bpi^l(\x_i)$ depends on the input and variant numbers of elements are larger than the threshold $\xi^l$. A series of sub-networks with various computational cost are obtained, which are intuitively  visualized  in Figure~\ref{fig-visual} of Section~\ref{sec-abl}.

\begin{table*}[t] 
	\centering
	\small 
	\caption{Comparison of the pruned ResNet with different methods on ImageNet~(ILSVRC-2012). `Top-1 Gap'/`Top-5 Gap' denotes the gaps of errors between the pruned models and the baseline models. `FLOPs $\downarrow$' is the reduction ratio of FLOPs.} 
	\begin{tabular}{|c|c|c|c|c|c|c|c|}
		\hline

		{Model} &   {Method}  & {Dynamic} & {Top-1 Error (\%)} &Top-5 Error (\%)&Top-1 Gap~(\%) &Top-5 Gap~(\%)  &  FLOPs $\downarrow$ (\%) \\
		
		\hline\hline
		
		\multirow{11}{*}{ResNet-18}  
		& Baseline& - & 30.24&10.92& 0.0 & 0.0&  0.0 \\
		& MIL~\cite{dong2017more} & \xmark & 33.67 & 13.06 & 3.43 & 2.14 &  33.3 \\
		& SFP~\cite{he2018soft} &\xmark  & 32.90   & 12.22 & 2.66 &1.30&  41.8 \\
		& FPGM~\cite{he2019filter} &\xmark& 31.59   & 11.52 & 1.35 & 0.60& 41.8 \\
			
		& PFP~\cite{Liebenwein2020Provable} & \xmark& 34.35  & 13.25 & 4.11 &2.33& 43.1 \\
		
		& DSA~\cite{ning2020dsa} &\xmark & 31.39 & 11.65 & 1.15 &0.73& 40.0 \\

		&LCCN~\cite{dong2017more}  & \cmark &33.67&13.06&3.43&2.14&34.6 \\ %
		&CGNet~\cite{hua2019channel}  & \cmark & 31.70 &-&1.46&-& 50.7 \\ %
		&FBS~\cite{gao2018dynamic}  & \cmark & 31.83 &11.78&1.59&0.86& 49.5 \\ %
		&ManiDP-A & \cmark &\textbf{31.12} &\textbf{11.24}&\textbf{0.88}&\textbf{0.32}&51.0 \\  
		&ManiDP-B  & \cmark & 31.65 &11.71&1.41&0.79&\textbf{55.1} \\ 
		\hline\hline 
		\multirow{10}{*}{ResNet-34} 
		& Baseline& - & 26.69&8.58& 0.0 &0.0 &  0.0 \\
		& SFP~\cite{he2018soft} &\xmark&  28.17   & 9.67  & 1.48 & 1.09& 41.1 \\
		& FPGM~\cite{he2019filter} & \xmark& 27.46    & 8.87 & 0.07 &0.29 &41.1 \\ 
		& Taylor~\cite{molchanov2019importance} & \xmark & 27.17   & -   & 0.48&-& 24.2 \\
		& DMC~\cite{gao2020discrete} & \xmark &27.43&8.89&0.74&0.31&43.4\\
		&LCCN~\cite{dong2017more}  & \cmark &27.01&8.81&0.32& 0.23&24.8 \\ %
		&CGNet~\cite{hua2019channel}  & \cmark & 28.70 &-&2.01&-& 50.4 \\ %
		&FBS~\cite{gao2018dynamic}  & \cmark &28.34&9.87&1.85&1.29& 51.2\\ 
		&ManiDP-A  & \cmark &\textbf{26.70}&\textbf{8.58}&\textbf{0.01}&\textbf{0.0}&46.8 \\ 
		&ManiDP-B & \cmark &27.26&8.96&0.57&0.38&\textbf{55.3} \\ 
		
		\hline
	\end{tabular}%
	\label{tab-resnet}%
\end{table*}%

\begin{table*}[t] 
	\centering
	\small 
	\caption{Comparison of the pruned MobileNetV2 with different methods on ImageNet~(ILSVRC-2012). `Top-1 Gap'/`Top-5 Gap' denotes the gaps of errors between the pruned models and the baseline models. `FLOPs $\downarrow$' is the reduction ratio of FLOPs.} 
	\begin{tabular}{|c|c|c|c|c|c|c|c|}
		\hline
			{Model} &   {Method}  & {Dynamic} & {Top-1 Error (\%)} &Top-5 Error (\%)&Top-1 Gap~(\%) &Top-5 Gap~(\%)  &  FLOPs $\downarrow$ (\%) \\
		
		\hline\hline
		
		\multirow{9}{*}{MobileNetV2}  
		& Baseline&- & 28.20&9.57&  0.0  &  0.0&0.0 \\  
		
		&ThiNet~\cite{luo2018thinet}&\xmark&36.25&14.59&8.05&5.02&44.7\\	
		& DCP~\cite{zhuang2018discrimination} & \xmark &35.78&-&7.58&-&44.7\\ 
		&MetaP~\cite{liu2019metapruning} &\xmark& 28.80&-&0.60&-&27.7\\
		& DMC~\cite{gao2020discrete} & \xmark &31.63&11.54&3.43&1.97&46.0\\
		&FBS~\cite{gao2018dynamic}  & \cmark & 29.07&9.91&0.87&0.34&33.6\\
		&ManiDP-A & \cmark & \textbf{28.58} &\textbf{9.72}&\textbf{0.38}&\textbf{0.15}&37.2 \\ 
		&ManiDP-B  & \cmark &30.38 &10.55&2.18&0.98&\textbf{51.2}\\ 
		
		\hline
	\end{tabular}%
	\label{tab-mbnet}%
	\vspace{-3mm}
\end{table*}%

\section{Experiments}
\label{sec-exp}
In this section, the proposed dynamic pruning method based manifold regularization~(ManiDP)  is empirically investigated on image classification datasets CIFAR-10~\cite{krizhevsky2009learning} and  ImageNet~(LSVRC-2012)~\cite{deng2009imagenet}. CIFAR-10 contains 60k 32$\times$32 colored images  from 10 categories, where  50k images are used as the training set and 10k for testing.  The large-scale ImageNet~(LSVRC-2012) dataset composes of 1.28M training images and 50k validation images, which are collected from 1k categories. Prevalent ResNet~\cite{he2016deep} models with different depths and light-weight MobilenetV2~\cite{sandler2018mobilenetv2} are used to verify the effectiveness of the proposed method.

\textbf{Implementation Details.}  For a fair comparison, the pruning rates for all layers in the network are the same following \cite{gao2018dynamic}. In the training phase, we increase the pruning rate from 0 to an appointed value $\xi$ to gradually make the pre-trained networks sparse. 
The coefficient $\lambda'$  regulating the weights of sparsity loss  is set to 0.005 for CIFAR-10 and 0.03 for ImageNet, empirically. The coefficient $\gamma$ for the similarity loss is set to 10 for both  two datasets.
All the networks are trained using the stochastic gradient descent(SGD)  with momentum 0.9. For CIFAR-10, the initial learning rate, batch-size and training epochs are set to 0.2, 128 and 300, respectively, while they are 0.25, 1024 and 120 for ImageNet. Standard data augmentation strategies containing random crop and horizontal flipping are used. For CIFAR-10, the images are padded to size 40$\times$40 and then cropped to size 32$\times$32. For ImageNet, images with resolution $224\times224$ are sent to the networks.  All the experiments are conducted with PyTorch~\cite{paszke2017automatic} on NVIDIA V100 GPUs.

\subsection{Comparison on ImageNet}
The proposed method is compared with state-of-the-art network pruning algorithms on the large-scale ImageNet dataset.
The pruning results of ResNet and MobileNetV2 are shown in Table~\ref{tab-resnet} and Table~\ref{tab-mbnet}, respectively, where the top-1/top-5 errors of  the pruned networks and the reduction ratios of FLOPs are reported. For dynamic pruning methods, the average FLOPs of the sub-networks over the whole test dataset are calculated as computational cost.

For  ResNet in Table~\ref{tab-resnet}, `ManiDP-A' and `ManiDP-B'  denote two pruned networks with different pruning rates, respectively.
The competing methods include both SOTA static channel pruning method developed recently~(\cite{he2018soft,he2019filter,ning2020dsa,li2020group,li2020dhp}) and the pioneering dynamic methods~(\cite{gao2018dynamic,dong2017more,hua2019channel}), indicated by \xmark~and \cmark~in the table.  Our method can reduce substantial computational cost for a given network with negligible performance degradation. For example, the proposed `ManiDP-A' can reduce 46.8\% FLOPs ResNet-34 with only 0.01\% performance degradation. Compared with the SOTA pruning algorithms, our method obtains pruned networks with less computational cost but lower test errors. The static methods are  obviously inferior to ours, \eg, the SOTA method DSA~\cite{ning2020dsa} only reduces 40.0\% FLOPs and obtain a pruned network with  31.39\% top-1 error (ResNet-18), while the proposed `ManiDP-A' can achieve lower test error (31.12\%) with more FLOPs reduced~(51.0\%). Our method  also shows superiority to  the existing dynamic pruning methods, \eg, FBS~\cite{gao2018dynamic} achieves 31.83\% top-1 error with 49.5\%  FLOPs pruned, which is worse than our method. We can infer that the proposed ManiDP method can excavate the redundancy of networks adequately to get compact but powerful networks with high performance.

\begin{table}[t] 
	\small			
	\caption{Realistic acceleration~(`Realistic Acl.') and theoretical acceleration (`Theoretical Acl.')  of pruned Networks on ImageNet.} 
	\label{tab-real}	
	\centering		
	\begin{tabular}{|c|c|c|c|}
		\hline 
		\multirow{2}{*}{Model}&\multirow{2}{*}{Method}&Theoretical&Realistic \\ 
		&&Acl.~(\%)& Acl.~(\%)\\ \hline \hline
		\multirow{2}{*}{ResNet-18}
		&ManiDP-A&51.0&35.4\\
		&ManiDP-B&55.6&40.5\\ \hline \hline
		\multirow{2}{*}{ResNet-34}
		&ManiDP-A &46.8&32.0\\
		&ManiDP-B&55.3&37.4 \\ \hline 
		\multirow{2}{*}{MobileNetV2}
		&ManiDP-A &37.2&30.4\\ 
		&ManiDP-B&51.2&38.5 \\ \hline 
		
	\end{tabular}
	\vspace{-3mm}
\end{table}

To validate the effectiveness of the proposed ManiDP method on light-weight networks, we  further compare it with SOTA methods on the efficient MobileNetV2~\cite{sandler2018mobilenetv2} designed for resource-limited devices, and the results are shown in Table~\ref{tab-mbnet}. Our method also achieves  a better trade-off between network accuracy and computational cost than the existing methods. For examples, the proposed `ManiDP-A' reduces 37.2\% FLOPs of MobileNetV2 with only 0.38\% accuracy loss, while the pruned network obtained by the competing method FBS~\cite{gao2018dynamic} sacrifices 0.87\% accuracy for pruning 33.6\% FLOPs. The results show that even light-weight networks are  over parameterized when exploring redundancy for each instances separately, which can be further accelerated by the proposed method and deployed on edge devices.

The realistic accelerations of the pruned Networks on ImageNet are shown in Table~\ref{tab-real}, which is calculated by counting the average inference time for handling each image on CPUs.  The realistic acceleration is slightly less than the theoretical acceleration calculated by FLOPs, which is due to practical factors such as I/O operations (\eg, accessing weights of networks), BLAS libraries and buffer switch, whose impact can be further reduced by practical engineering optimization.

\begin{table}[t] 
	\centering
	\small 
	\caption{Comparison of the pruned ResNet with different methods on CIFAR-10. } 
	\begin{tabular}{|c|c|c|c|c|}
		\hline
		{Depth} &   {Method}  & {Dynamic} & {Error (\%)} &  FLOPs $\downarrow$ (\%) \\
	
		\hline\hline
		
		\multirow{8}{*}{20}  
		& Baseline& -& 7.78 & 0.0  \\
		& SFP\cite{he2018soft}& \xmark&9.17      & 42.2 \\
		& FPGM \cite{he2019filter} & \xmark & 9.56 &      54.0 \\  
		& DSA~\cite{ning2020dsa} & \xmark& 8.62    & 50.3 \\
		&Hinge~\cite{li2020group}&\xmark&8.16&45.5\\
		& DHP~\cite{li2020dhp} &\xmark&8.46&51.8 \\
		&FBS~\cite{gao2018dynamic}   & \cmark &9.03& 53.1  \\
		&ManiDP  & \cmark &\textbf{7.95}&\textbf{54.2} \\ 
		\hline \hline
		\multirow{6}{*}{32}  
		& Baseline  &-& 7.34& 0.0\\  
		& MIL\cite{dong2017more}& \xmark & 9.26 &   31.2 \\
		& SFP \cite{he2018soft} & \xmark& 7.92 &   41.5 \\
		& FPGM \cite{he2019filter} & \xmark& 8.07     &53.2 \\
	
		&FBS~\cite{gao2018dynamic}   & \cmark &8.02&  55.7  \\  
		&ManiDP  & \cmark &\textbf{7.85}& \textbf{63.2}   \\ 
		\hline \hline
		
		\multirow{9}{*}{56} 
		& Baseline  &-& 6.30&0.0\\
		&  SFP \cite{he2018soft}   & \xmark & 7.74  &        52.6 \\
		&  FPGM \cite{he2019filter}    &  \xmark & 6.51  &   52.6 \\ 
		
		&   HRank~\cite{lin2020hrank}    &  \xmark  &  6.83      &        50.0\\
		& DSA~\cite{ning2020dsa} &\xmark & 7.09    & 52.2 \\
		&Hinge~\cite{li2020group}&\xmark&6.31&50.0\\
		& DHP~\cite{li2020dhp} &\xmark&6.42&50.9 \\
		&FBS~\cite{gao2018dynamic}   & \cmark &6.48& 53.6  \\ 
		&ManiDP  & \cmark &\textbf{6.36}&\textbf{62.4} \\ \hline  

	\end{tabular}
	\label{tab-cifar}
\end{table}%

\begin{table}[t]
	\centering
	\small	
	\caption{Effectiveness of  Excavating Manifold Information. The top-1 errors of the pruned networks and the gaps from the base networks are reported.} 
	\begin{tabular}{|c|c|c|c|}
		\hline
		{Model} & {Complexity} & {Similarity}&Error / Gap~(\%) \\
		\hline \hline

		&\xmark&\xmark &  6.88 /  0.58  \\ 
		ResNet-56&  \cmark &\xmark& 6.61 /  0.31  \\
		(CIFAR-10)&\xmark   & \cmark & 6.53 /  0.23\\  
		&\cmark & \cmark & \textbf{6.36} /  \textbf{0.06} \\  \hline\hline
		&\xmark&\xmark &  28.12 /  1.43 \\
		ResNet-34&  \cmark &\xmark&  27.67 /  0.98 \\
		(ImageNet)&\xmark   & \cmark &27.63 /   0.94 \\  
		&\cmark & \cmark & \textbf{27.29} /  \textbf{0.57} \\
		\hline 
	\end{tabular}
	\label{tab-mani}
	\vspace{-3mm}
\end{table}%

\subsection{Comparison on CIFAR-10}
On the benchmark CIFAR-10 dataset, the comparison between the proposed ManiDP and SOTA channel pruning  methods are shown in Table~\ref{tab-cifar}. Compared with  SOTA methods, a significantly higher FLOPs reduction is achieved by our method with less degradation of performance. For example, using our method, more than 60\% FLOPs of the ResNet-56 model are reduced while the test error can still achieve 6.36\% using ManiDP. Compared to the static methods (\eg,~HRank~\cite{lin2020hrank} with 6.83\% error and 50.0\% FLOPs reduction) and dynamic method (\eg, FBS~\cite{gao2018dynamic} with 6.48\% error and 53.6\% FLOPs reduction), our method shows notable superiority.

\subsection{Ablation Studies}
\label{sec-abl} 

\textbf{Effectiveness of Manifold Information.} To maximally excavate network redundancy corresponding to  each instance, the manifold information between instances is explored from two perspectives, \ie, complexity and similarity. The impacts of whether exploiting complexity or similarity relationship is empirically investigated in Table~\ref{tab-mani}, indicated by \cmark~and \xmark. The classification error and the performance gap compared to the base models are reported. 
Without utilizing the complexity relationship means fixing the trade-off coefficient $\lambda(\x_i)$ between lasso loss and cross-entropy loss, which obviously increases the error incurred by pruning~(\eg, 1.43\% \vs 0.94\% on ImageNet). The unsatisfactory performance is due to the improper alignment, \ie, cumbersome sub-networks may be assigned to simple examples, while complex instances are handled by tiny sub-networks with limited representation capability.  For the similarity relationship, deactivating it (setting coefficient $\gamma$ for similarity loss to zero) incurs larger performance degradation~(\eg, 1.43\% \vs 0.98\%), which validates the effectiveness of exploring the similarity between features  of instances and the corrsponding sub-networks.  Thus, exploiting both the two perspectives of manifold information is necessary to achieve negligible performance degradation (\ie, only 0.57\% error increase on ImageNet).

 \begin{figure}[t] 
 	\small
	\subfigure[]{
		\begin{minipage}[t]{0.47\linewidth}
			\centering
			\includegraphics[width=0.99\linewidth]{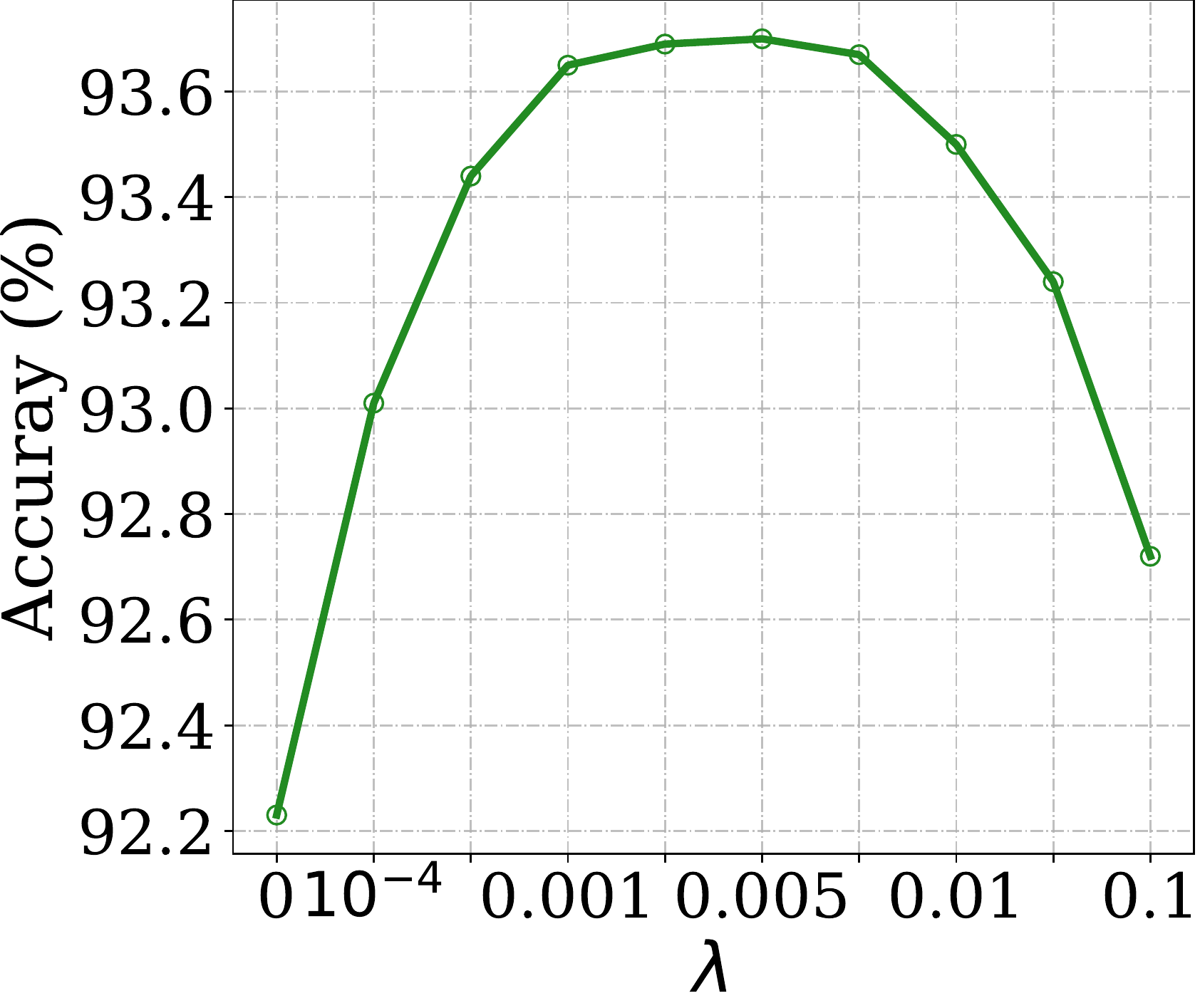}
		\end{minipage}
	}
	\subfigure[]{
		\begin{minipage}[t]{0.47\linewidth}
			\centering
			\includegraphics[width=0.99\linewidth]{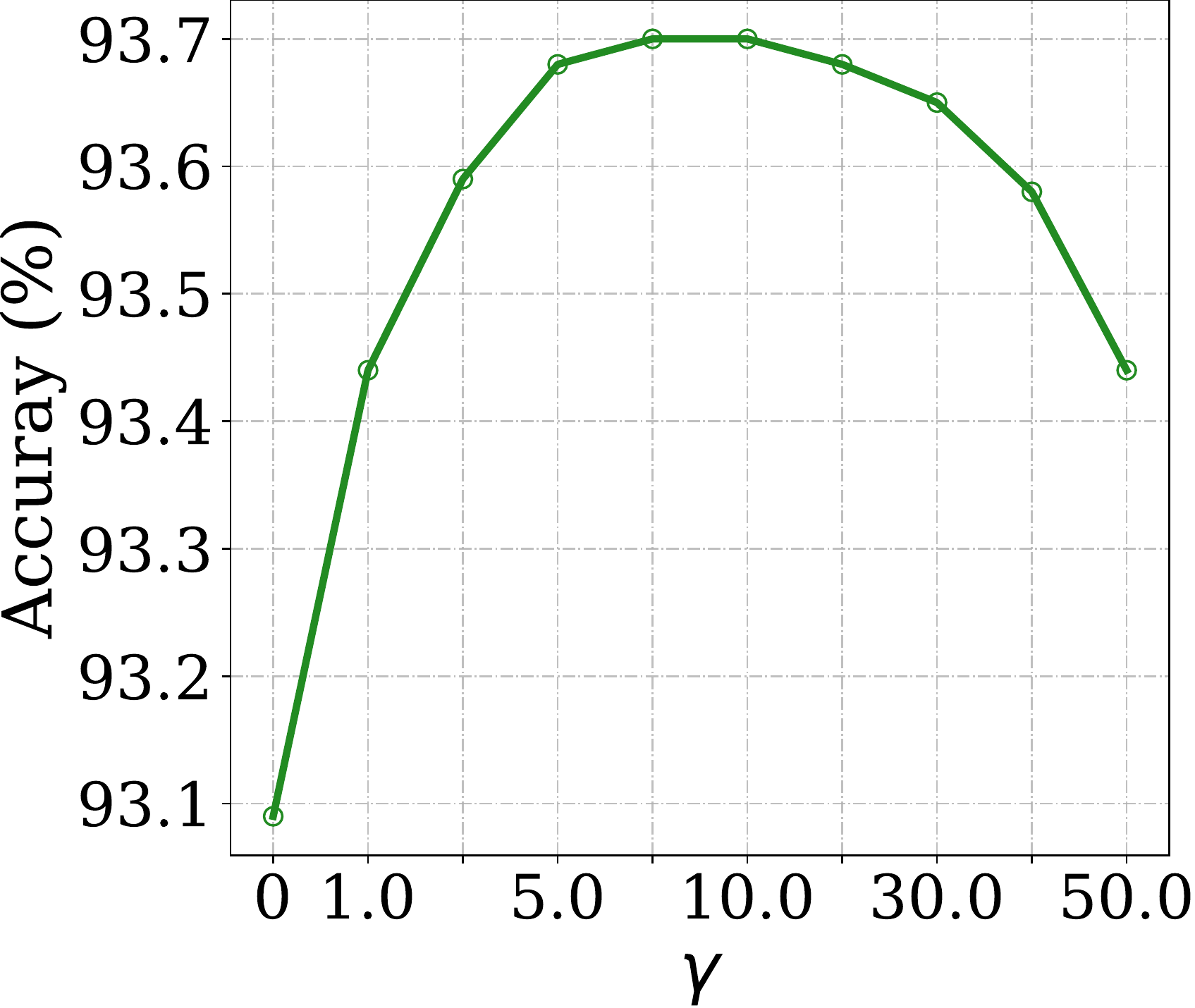}
		\end{minipage}
	}

	\caption{Test accuracies of the pruned ResNet-56 \wrt (a) weight coefficient $\lambda'$ for sparsity loss and (b) coefficient $\mu$ for similarity loss.}
	\label{fig-hyper}
\end{figure}

\begin{figure}[t]
	\vspace{-1mm}
	\small
	\centering
	\includegraphics[width=0.63\columnwidth]{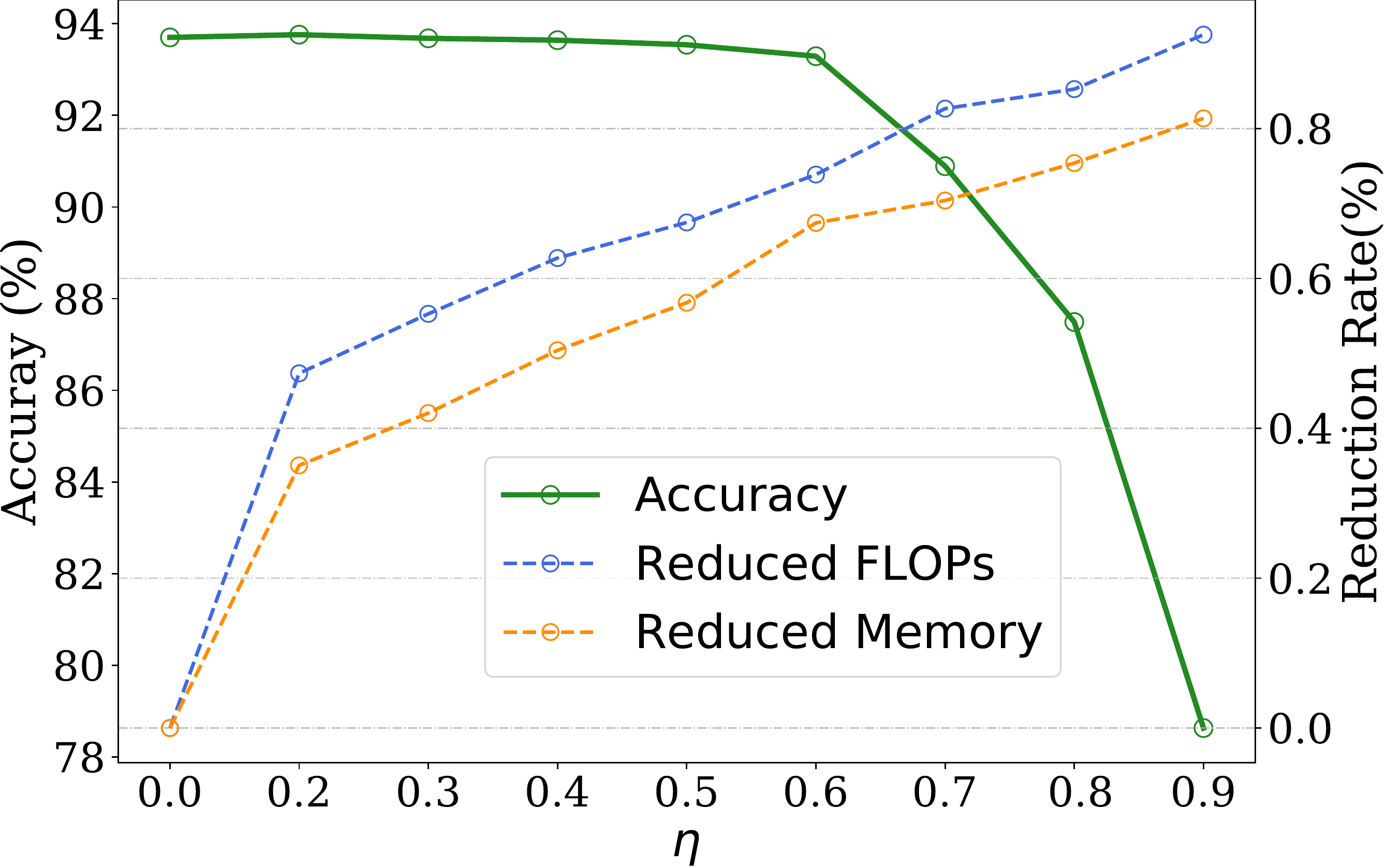} %
	\caption{The variety of test accuracies and required Memory \& FLOPs (ResNet-56)  \wrt pruning rate $\eta$.}
	\label{fig-pru} 
	\vspace{-5mm}
\end{figure}

\textbf{Weight coefficients $\lambda'$ and $\gamma$}. The weights of sparsity loss  and similarity loss are controlled by coefficients $\lambda'$~(Eq.~(\ref{eq-lasso2})) and $\gamma$~(Eq.~\ref{eq-final}), whose impact on the final test accuracies is shown in Figure~\ref{fig-hyper}. A larger $\lambda'$ induces more sparsity on channel saliencies, which will have less impact on the network outputs when discarding channels with small saliencies. On the other hand, the sparsity will affect the representation ability of networks and incur accuracy drop~(Figure~\ref{fig-hyper}~(a)). Analogous phenomenon exists when varying coefficient $\gamma$ for similarity loss. The test accuracy of the pruned network is improved when increasing $\gamma$ unless it is set to an extremely large value, as the similarity between different instances is excavated more adequately.  Note that our method is robust to both hyper-parameters and works well in a wide range (\eg, range [0.001,0.01] for $\lambda$ and values around 10.0 for $\gamma$), empirically.

\textbf{Memory/FLOPs and accuracies \wrt Pruning Rate.} The impact of different pruning rate is shown in Figure~\ref{fig-pru}. When a single instance is sent to the network, the memory cost for accessing network weights can be reduced as ineffective weights do not participate the inference process. With a large reduction of computational cost and  memory, the pruned network can still achieve a high performance. For example, when setting the pruning rare to 0.6\% with 73.87\% FlOPs and 67.42\% memory reduction, the pruned ResNet-56 can still achieve an accuracy of 93.29\%~(only 0.41\% accuracy drop compared to the original network).

\textbf{Visualization.}
Different sub-networks with various computational costs (\ie, FLOPs) are generated for each instance by pruning different channels. Using ResNet-34 as the backbone, the  FLOPs distribution  of different sub-networks  over the validation set of ImageNet are shown in Figure~\ref{fig-visual}, where $x$-axis denotes FLOPs and the $y$-axis is  the number of sub-networks. The FLOPs of sub-networks varies in a certain range  \wrt the complexity of instances. Most of the sub-networks own medium sizes and a small quantity of sub-networks activate more/less channels to handle harder/simpler instances. Some representative images handled by the corresponding sub-networks are also shown in the figure.  
Intuitively, a simple example (\eg, `bird' and `dog' in the red frames) that can be correctly predicted by a compact network usually contains clear targets, while images with obscure semantic information (\eg, too large `orange' and too small `flower' in the blue frames) require larger networks with more powerful representation ability. More visualization results are shown in the supplementary material.

\begin{figure}[t]
	\centering
	\includegraphics[width=0.7\columnwidth]{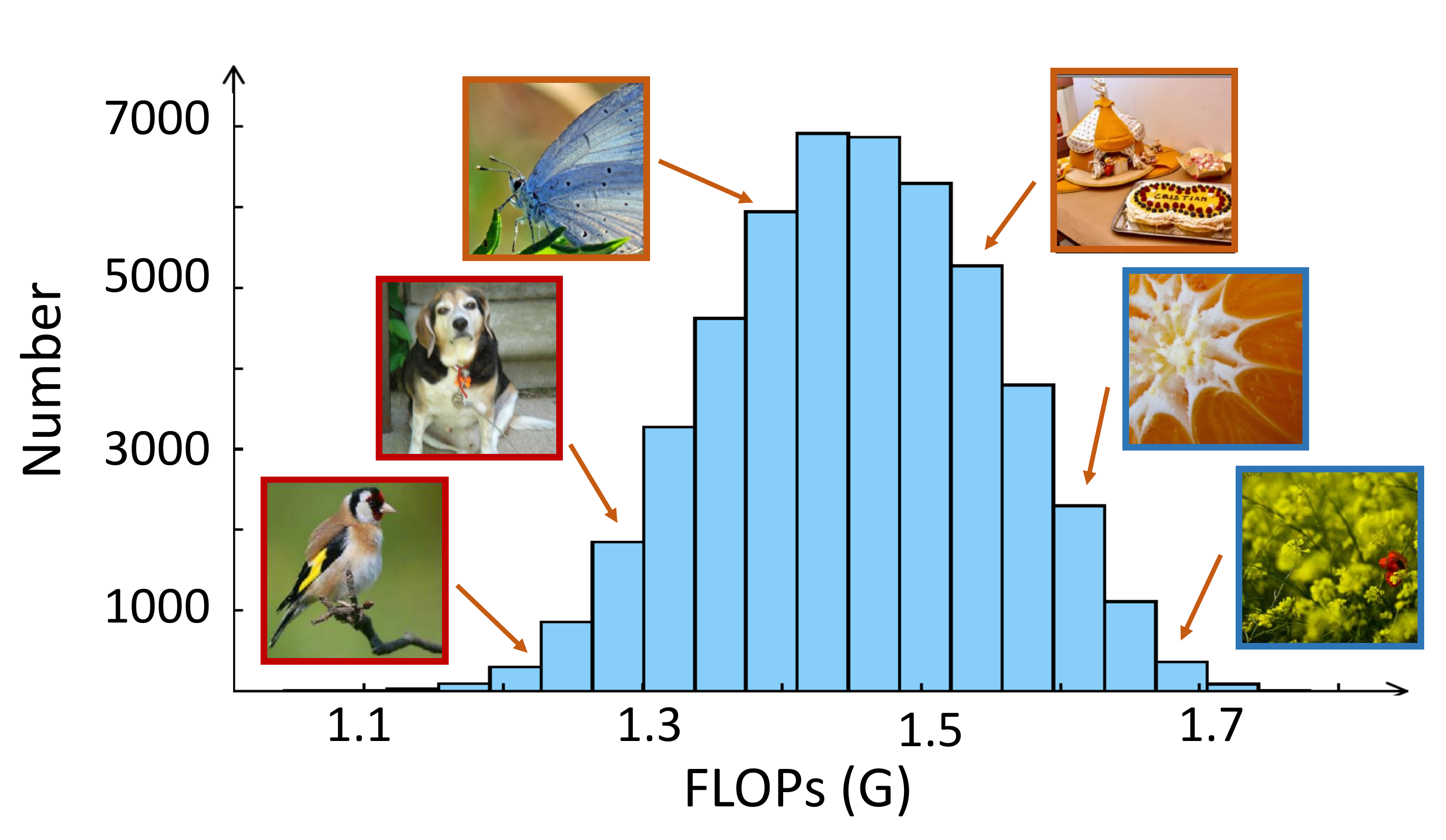} %
	\caption{FLOPs distribution of sub-networks and their corresponding input instances on ImageNet.}
	\label{fig-visual} 
	\vspace{-5mm}
\end{figure}

\vspace{-1mm}
\section{Conclusion}
\vspace{-1mm}
\label{sec-con}
This paper proposes a manifold regularized dynamic pruning method (ManiDP) to maximally excavate the redundancy of neural networks. We explore the manifold information in the sample space to discover the relationship between different instances from two perspectives, \ie, complexity and similarity, and then the relationship is preserved in the corresponding sub-networks. An adaptive penalty weight for network sparsity  is developed to align  the instance complexity and network complexity, while the  similarity relationship is preserved by matching the similarity matrices. Extensive experiments are conducted on several benchmarks to verify the  effectiveness of our method. Compared with the state-of-the-art methods,  the pruned networks obtained by the proposed ManiDP can achieve better performance with less computational cost.  For example, our method can reduce 55.3\% FLOPs of ResNet-34 with only 0.57\% top-1 accuracy degradation on ImageNet.

\noindent\textbf{Acknowledgment.} This work is supported by National Natural Science Foundation of China under Grant No. 61876007, and Australian Research Council under Project DE180101438 and DP210101859.

{\small
\bibliographystyle{ieee_fullname}
\bibliography{egbib}
}

\clearpage

\begin{figure}[t]
	\centering
	\includegraphics[width=1.0\columnwidth]{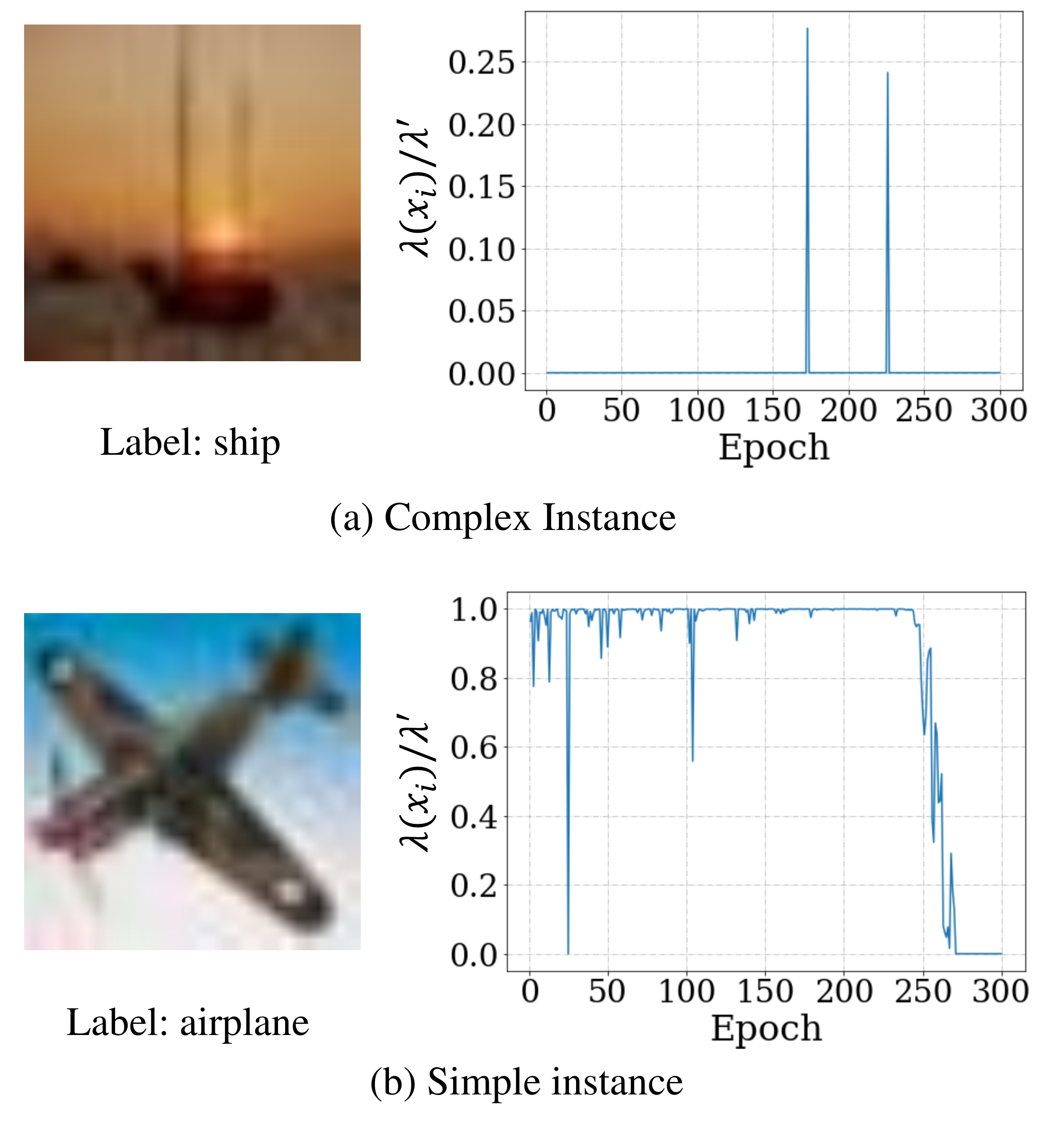} %
	\caption{Coefficient $\lambda(\x_i)$ that controls network sparsity for complex and simple instances.}
	\label{sfig-lambda}
	
\end{figure}

\section{Supplementary Material}

\subsection{Coefficient $\lambda(\x_i)$ for different instances}

In the training procedure, the  coefficient $\lambda(\x_i)$ (Eq.~(6) of the main paper) controls the weight of sparsity loss according to the complexity of each instance. Recall that $\lambda(\x_i)=\lambda' \cdot \beta_i\frac{C-\L_{ce}(\x_i,\W)}{C} \in [0,\lambda']$ , where $\lambda'$ is a fixed hyper-parameter for all instances. Using ResNet-56 as the backbone, the variable parts $\lambda(\x_i)/\lambda' \in [0,1]$ for  complex/simple examples in CIFAR-10 are shown in Figure~\ref{sfig-lambda}.  $\lambda(\x_i)/\lambda'$ keeps small for complex examples (\eg, the vague `ship' in  (a)) and then less channels of the pre-defined networks are pruned for keeping their representation capabilities.  When sending simple examples (\eg, clear `airplane' in (b)) to the dynamic network, $\lambda(\x_i)/\lambda'$ keeps large in most of the epochs, and thus the corresponding sub-network becomes sparser  continuously as the numbers of iteration increases. Note that $\lambda(\x_i)/\lambda'$  changes dynamically in the training process. For example, the sparsity weight automatically decreases in the last few epochs as the corresponding sub-network is compact enough and should pay more attention to accuracy (Figure~\ref{sfig-lambda}~(b)), which ensures that the models can fit input instances well.



\begin{figure}[t]
	\centering
	\includegraphics[width=1.0\columnwidth]{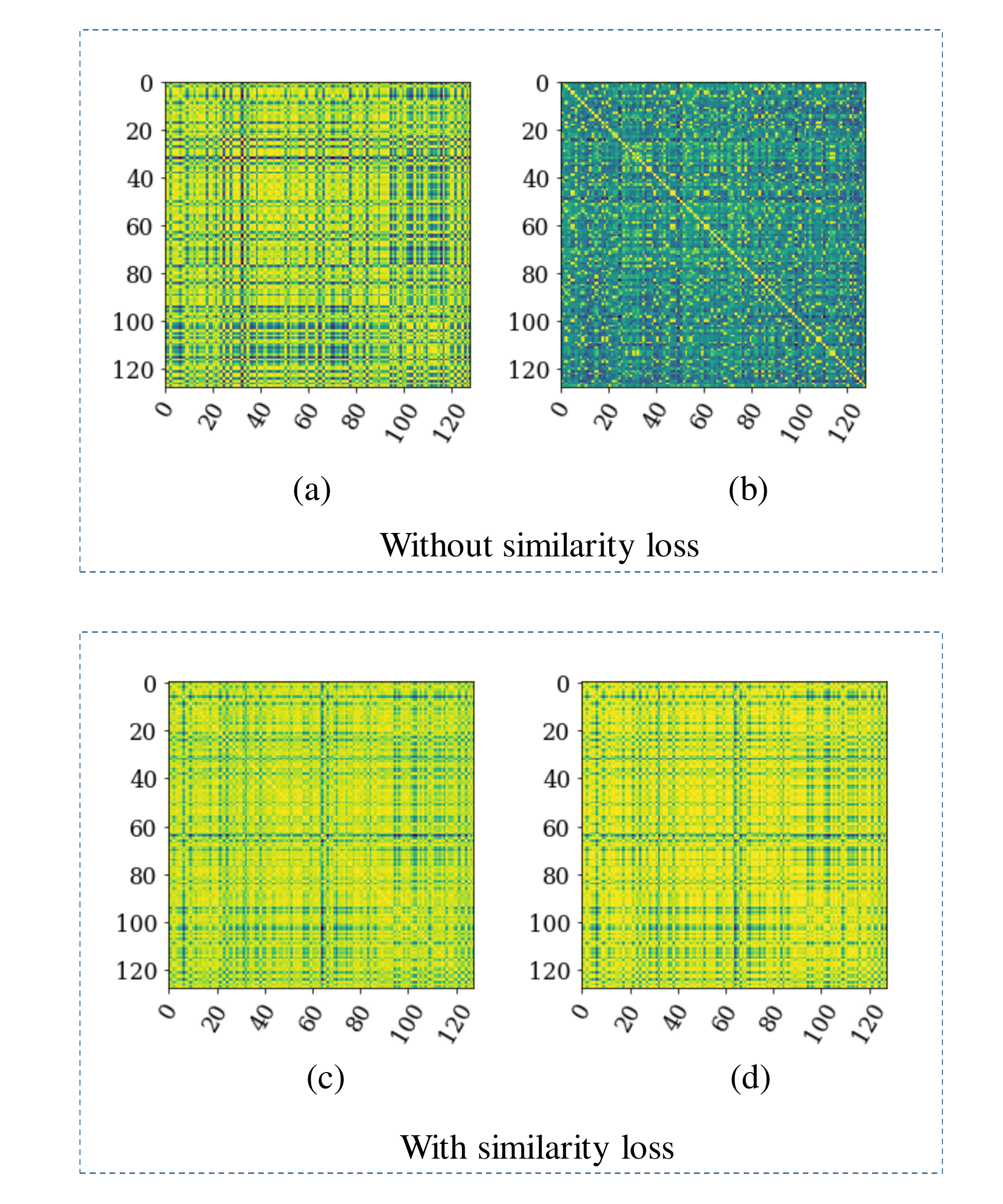} %
	\caption{Similarity matrices. (a) Similarity matrix $R^l$ for intermediate features  without similarity loss. (b) Similarity matrix $T^l$ for channel
		saliencies without similarity loss. (c) Similarity matrix $R^l$ for intermediate features  with similarity loss. (d) Similarity matrix $T^l$ for channel
		saliencies with similarity loss.}
	\label{sfig-sim} 
	
\end{figure}
\subsection{Similarity Matrices}
The similarity matrices $R^l$ for intermediate features and $T^l$ for channel
saliencies are shown in Figure~\ref{sfig-sim}, where different colors denote the degree of similarity (\ie, a yellower point means higher degree of similarity between  two instances). The ResNet-56 model trained  with/without the similarity loss $\L_{sim}$ (Eq.~(10) in the main paper) is used to generate features and channel saliencies for calculating the similarity between instances randomly sampled from CIFAR-10. When training dynamic network without similarity loss $\L_{sim}$, the similarity calculated by features and that by channel saliencies are very  different (Figure~\ref{sfig-sim}~(a),~(b)). When using similarity loss (Figure~\ref{sfig-sim}~(c),~(d)), the  similarity matrices $R^l$ and $T^l$ are more analogous as the  similarity loss penalizes the inconsistency between  similarity matrices  to align the similarity relationship in the two spaces.

\subsection{Visualization of instances with different complexity}

We sample representative images with different complexity from ImageNet and intuitively show them in Figure~\ref{fig-visual}.  From top to bottom, the computational costs of sub-networks used to predict labels continue to increase. Intuitively, simple instances that can be accurately predicted by compacted networks usually  contain clear targets, while the semantic information in complex images are vague and thus requires larger networks  with powerful representation capability.

\begin{figure}[h]
	\centering
	\includegraphics[width=1.0\columnwidth]{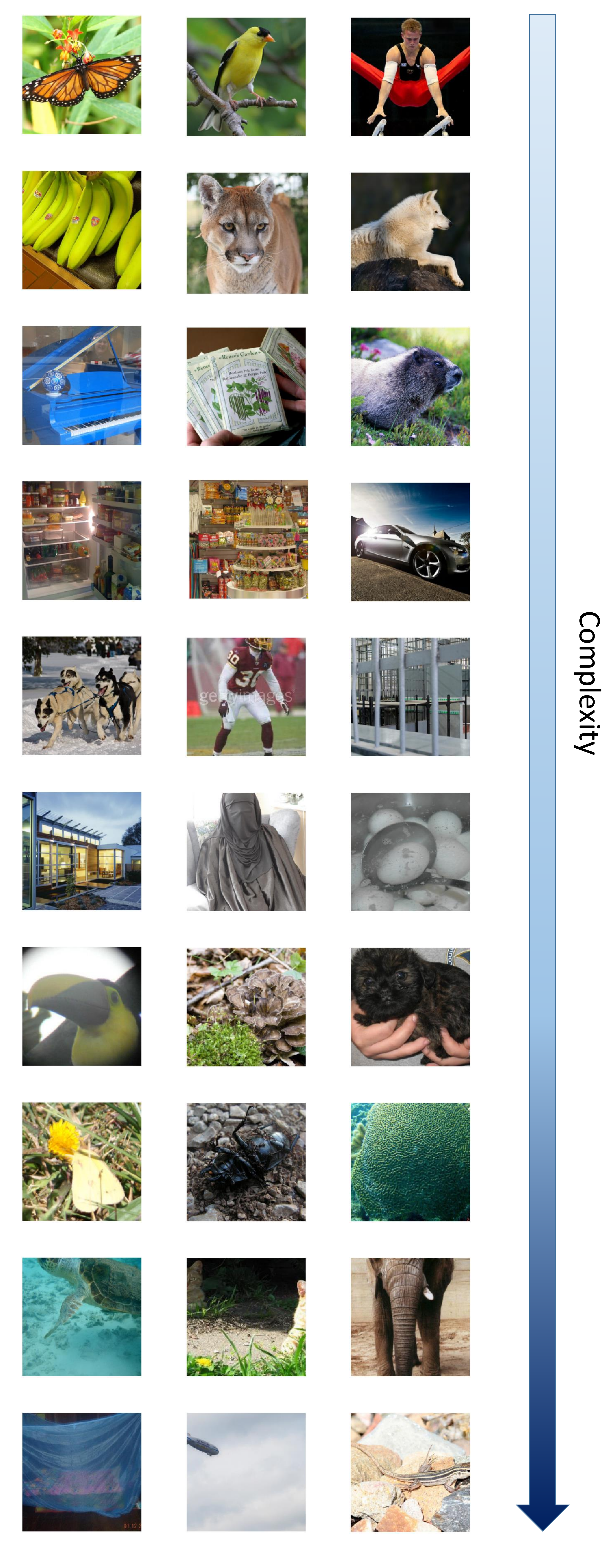} %
	\caption{ Images with different complexity on ImageNet.  From top to bottom, the computational costs of sub-networks used to predict labels continue to increase.}
	\label{fig-visual} 
\end{figure}

\end{document}